\documentclass{amsart}
\usepackage{amsaddr}
\usepackage{hyperref}
\usepackage{graphicx}

\begin{document}

\title{Evolving Evolutionary Algorithms using \\Linear Genetic Programming}

\author{Mihai Oltean}

\address{Department of Computer Science,\\
Faculty of Mathematics and Computer Science,\\
Babes-Bolyai University, Kogalniceanu 1,\\
Cluj-Napoca, 3400, Romania.\\
\url{https://mihaioltean.github.io}
}
\email{mihai.oltean@gmail.com}

\maketitle

\begin{abstract}

A new model for evolving Evolutionary Algorithms is proposed in this paper. The model is based on the Linear Genetic Programming (LGP) technique. Every LGP chromosome encodes an EA which is used for solving a particular problem. Several Evolutionary Algorithms for function optimization, the Traveling Salesman Problem and the Quadratic Assignment Problem are evolved by using the considered model. Numerical experiments show that the evolved Evolutionary Algorithms perform similarly and sometimes even better than standard approaches for several well-known benchmarking problems.

\end{abstract}

\section{Introduction}

Evolutionary Algorithms (EAs) (Goldberg, 1989; Holland, 1975) are new and powerful tools used for solving 
difficult real-world problems. They have been developed in order to solve some real-world problems that the classical (mathematical) methods failed to successfully tackle. Many of these unsolved problems 
are (or could be turned into) optimization problems. The solving of an optimization 
problem means finding solutions that maximize or minimize a criteria 
function (Goldberg, 1989; Holland, 1975; Yao et al., 1999).

Many Evolutionary Algorithms have been proposed for dealing with optimization problems. Many 
solution representations and search operators have been proposed and tested 
within a wide range of evolutionary models. There are several natural 
questions to be answered in all these evolutionary models: 

\textit{What is the optimal population size?}

\textit{What is the optimal individual representation?}

\textit{What are the optimal probabilities for applying specific genetic operators?}

\textit{What is the optimal number of generations before halting the evolution?}

A breakthrough arose in 1995 when Wolpert and McReady unveiled their 
work on \textit{No Free Lunch} (NFL) theorems for \textit{Search} (Wolpert et al., 1995) and \textit{Optimization} (Wolpert et al., 1997). The No Free Lunch theorems state that all the black-box algorithms have the same average performance over the entire set of optimization problems. (A black-box algorithm does not take into account any information 
about the problem or the particular instance being solved.) The magnitude of the 
NFL results stroke all the efforts for developing a universal 
black-box optimization algorithm capable of solving all the optimization 
problems in the best manner. Since we cannot build an EA able to solve best all problems we have to find other ways to construct algorithms that perform very well for some particular problems. One possibility (explored in this paper) is to let the evolution to discover the optimal structure and parameters for the evolutionary algorithm used for solving a particular problem.

In their attempt for solving problems, men delegated computers to develop 
algorithms capable of performing certain tasks. The most prominent effort in this 
direction is Genetic Programming (GP) (Koza, 1992; Koza, 1994), an evolutionary technique 
used for breeding a population of computer programs. Instead of evolving 
solutions for a particular problem instance, GP is mainly intended for 
discovering computer programs capable of solving particular classes of 
optimization problems. (This statement is only partially true since the 
discovery of computer programs may also be viewed as a technique for 
solving a particular problem input. For instance, the problem may be 
here: "Find a computer program that calculates the sum of the elements of 
an array of integers.").

There are many such approaches in literature concerning GP. Noticeable effort has been 
dedicated for evolving deterministic computer programs capable of solving specific 
problems such as symbolic regression (Koza, 1992; Koza, 1994), classification (Brameier et al., 2001a) etc.

Instead of evolving such deterministic computer programs we will evolve a 
full-featured evolutionary algorithm (i.e. the output of our main program 
will be an EA capable of performing a given task). Thus, we will work with EAs at 
two levels: the first (macro) level consists in a steady-state EA (Syswerda, 1989) which 
uses a fixed population size, a fixed mutation probability, a fixed crossover 
probability etc. The second (micro) level consists in the solutions encoded 
in a chromosome of the first level EA.

For the first (macro) level EA we use an evolutionary model similar to 
Linear Genetic Programming (LGP) (Brameier et al., 2001a; Brameier et al., 2001b; Brameier et al., 2002) which is very suitable for evolving computer programs that may be easily translated into an imperative 
language (like \textbf{\textit{C}} or \textbf{\textit{Pascal}}).

The rules employed by the evolved EAs during of a generation are not preprogrammed. These rules are 
automatically discovered by the evolution. The evolved EA is a generational 
one (the generations do not overlap).

This research was motivated by the need of answering several important 
questions concerning Evolutionary Algorithms. The most important question 
is "Can Evolutionary Algorithms be automatically synthesized by using only 
the information about the problem being solved?" (Ross, 2002). And, if yes, which are 
the genetic operators that have to be used in conjunction with an EA (for a 
given problem)? Moreover, we are also interested to find the 
optimal (or near-optimal) sequence of genetic operations (selections, 
crossovers and mutations) to be performed during a generation of an 
Evolutionary Algorithm for a particular problem. For instance, in a standard 
GA the sequence is the following: selection, recombination and mutation. But, how do we know that scheme is the best for a 
particular problem (or problem instance)? We better let the evolution to 
find the answer for us.

Several attempts for evolving Evolutionary Algorithms were made in the past (Ross, 2002; Tavares et al., 2004). A 
non-generational EA was evolved (Oltean et al., 2003) by using the Multi Expression 
Programming (MEP) technique (Oltean et al., 2003; Oltean, 2003).

There are also several approaches that evolve genetic operators for solving 
difficult problems (Angeline, 1995; Angeline, 1996; Edmonds, 2001; Stephens et al., 1998; Teller, 1996). In his paper on Meta-Genetic Programming, Edmonds (Edmonds, 2001) used two populations: a standard GP population and a co-evolved population of operators that act on the main population. Note that all these approaches use a fixed evolutionary algorithm which is not changed during the search.

A recent paper of Spector and Robinson (Spector, 2002) describes a language called Push which supports a new, self-adaptive form of evolutionary computation called \textit{autoconstructive evolution}. An experiment for symbolic regression problems was reported. The conclusion was that "Under most conditions the population quickly achieves reproductive competence and soon thereafter improves in fitness." (Spector 2002).

There are also several attempts for evolving heuristics for particular problems. In (Oltean et al., 2004a) the authors evolve an heuristic for the Traveling Salesman Problem. The obtained heuristic is a mathematical expression that takes as input some information about the already constructed path and outputs the next node of the path. It was shown (Oltean et al., 2004a) that the evolved heuristic performs better than other well-known heuristics (Nearest Neighbor Heuristic, Minimum Spanning Tree Heuristic (Cormen et al., 1990; Garey et al., 1979)) for the considered test problems.

The paper is organized as follows. The LGP technique is 
described in section \ref{sec2}. The model used for evolving EAs is presented in section \ref{sec3}. 
Several numerical experiments are performed in section \ref{sec4}. Three EAs for 
function optimization, the Traveling Salesman Problem  and the Quadratic Assignment Problem are evolved in sections \ref{funcopt}, \ref{tsp} and \ref{qap}. Further research directions are suggested in section \ref{future}.

\section{Linear Genetic Programming Technique}\label{sec2}

In this section the \textit{Linear Genetic Programming} (LGP) technique is described. LGP uses a linear chromosome 
representation and a special phenotype transcription model.

\subsection{LGP Algorithm} \label{ss}

In our experiments steady-state (Syswerda, 1989) is used as underlying mechanism for LGP.

The steady-state LGP algorithm starts with a randomly chosen population of 
individuals. The following steps are repeated until a termination condition 
is reached: Two parents are selected by using binary tournament 
and are recombined with a fixed crossover probability. Two offspring are obtained by the recombination 
of two parents. The offspring are mutated and the 
best of them replaces the worst individual in the current population (if the 
offspring is better than the worst individual in the current population).

\subsection{Individual Representation}

\textit{Linear Genetic Programming} (LGP) (Banzhaf et al., 1998; Brameier et al., 2001a; Nordin, 1994) uses a specific linear representation of computer programs. Programs of an imperative language (like 
\textbf{\textit{C}}) are evolved instead of the tree-based GP expressions of a functional programming language 
(like \textbf{\textit{LISP}}). 

An LGP individual is represented by a variable-length sequence of simple 
\textbf{\textit{C}} language instructions. Instructions operate on one or 
two indexed variables (registers) $r$ or on constants $c$ from predefined sets. 
The result is assigned to a destination register, e.g. $r_{i}=r_{j}$ * 
$c$. 

An example of an LGP program is the following:\\

\textsf{\textbf{void}}\textsf{ 
LGP{\_}Program(}\textsf{\textbf{double}}\textsf{ v[8])}

\textsf{{\{}}

\textsf{\ldots }

\textsf{v[0] = v[5] + 73;}

\textsf{v[7] = v[4] - 59;}

\textsf{v[4] = v[2] *v[1];}

\textsf{v[2] = v[5] + v[4];}

\textsf{v[6] = v[1] * 25;}

\textsf{v[6] = v[4] - 4;}

\textsf{v[1] = sin(v[6]);}

\textsf{v[3] = v[5] * v[5];}

\textsf{v[7] = v[6] * 2;}

\textsf{v[5] = [7] + 115;}

\textsf{v[1] = sin(v[7]);}

\textsf{{\}}}\\

A linear genetic program can be turned into a functional representation by 
successive replacements of variables starting with the last effective 
instruction (Brameier et al., 2001a).

Variation operators are crossover and mutation. By crossover 
continuous sequences of instructions are selected and exchanged between 
parents (Brameier et al., 2001a). Two cutting points are randomly chosen in each parent and the sequences of instructions between them are exchanged. As an immediate effect, the length of the obtained offspring might be different from the parents.

Two types of mutations are used: micro mutation and macro mutation (Brameier et al., 2001a). 
By micro mutation an operand or an operator of an instruction is changed. 
Macro mutation inserts or deletes a random instruction.

\section{LGP for Evolving Evolutionary Algorithms}\label{sec3}

In order to use LGP for evolving EAs we have to modify the structure of an 
LGP chromosome and define a set of function symbols.

\subsection{Individual Representation for Evolving EAs}

Instead of working with registers, our LGP program will modify an array of 
individuals (the population). We denote by \textit{Pop} the array of 
individuals (the population) which will be modified by an LGP program. 

The set of function symbols will consist in genetic operators that may 
appear into an evolutionary algorithm. There are usually 3 types of genetic 
operators that may appear into an EA. These genetic operators are:\\

\textit{Select} - selects the best solution among several already existing solutions,\\

\textit{Crossover} - recombines two existing solutions,\\

\textit{Mutate} - varies an existing solution.\\

These operators will act as function symbols that may appear into 
an LGP chromosome. Thus, each simple \textbf{\textit{C}} instruction that 
appeared into a standard LGP chromosome will be replaced by a more complex 
instruction containing genetic operators. More specifically, we have three major types of instructions in the modified LGP 
chromosomes. These instructions are:\\

\textit{Pop}[$k$] = \textit{Select} (\textit{Pop}[$i$], \textit{Pop}[$j$]); \textsf{// Select the best individual from those stored in }

\hspace{4cm}\textsf{// }\textsf{\textit{Pop}}\textsf{[}\textsf{\textit{i}}\textsf{] and 
}\textsf{\textit{Pop}}\textsf{[}\textsf{\textit{j}}\textsf{] and keep the 
result in position }\textsf{\textit{k}}\textsf{.}\\

\textit{Pop}[$k$] = \textit{Crossover} (\textit{Pop}[$i$], \textit{Pop}[$j$]); \textsf{// Crossover the individuals stored in }

\hspace{4cm}\textsf{// }\textsf{\textit{Pop}}\textsf{[}\textsf{\textit{i}}\textsf{] and 
}\textsf{\textit{Pop}}\textsf{[}\textsf{\textit{j}}\textsf{] and keep the 
result in position }\textsf{\textit{k}}\textsf{.}\\

\textit{Pop}[$k$] = \textit{Mutate} (\textit{Pop}[$i$]); \textsf{// Mutate the individual stored in }

\hspace{4cm}\textsf{// position }\textsf{\textit{i}}\textsf{ and keep the result in 
position }\textsf{\textit{k}}\textsf{.}\\

\textit{Remarks}:

\begin{itemize}

\item[{\it (i)}]{The \textit{Crossover} operator always generates a single offspring from two parents in our model. 
Crossover operators generating two offspring may be designed to fit our 
evolutionary model as well.}

\item[{\it (ii)}]{The \textit{Select} operator acts as a binary tournament selection. The better of two 
individuals is always accepted as the result of the selection.}

\item[{\it (iii)}]{\textit{Crossover} and \textit{Mutate} operators are problem dependent. For instance, if we want to evolve an 
EA (with binary representation) for function optimization we may use the set 
of genetic operators having the following functionality: \textit{Crossover} -- recombines two 
parents using one cut point crossover, \textit{Mutate} -- one point mutation. If we want to 
evolve an EA for solving the TSP problem (Merz et al., 1997) we may use DPX as a crossover 
operator and 2-opt as a mutation operator (Krasnogor, 2002).}

\end{itemize}

An LGP chromosome $C$, storing an evolutionary algorithm is the following:\\

\textsf{\textbf{void}}\textsf{ 
LGP{\_}Program(}\textsf{\textbf{Chromosome}}
\textsf{ Pop[8]) // a population with 8 individuals}

\textsf{{\{}}

\textsf{...}

\hspace{0.5cm}\textsf{Pop[0] = Mutate(Pop[5]);}

\hspace{0.5cm}\textsf{Pop[7] = Select(Pop[3], Pop[6]);}

\hspace{0.5cm}\textsf{Pop[4] = Mutate(Pop[2]);}

\hspace{0.5cm}\textsf{Pop[2] = Crossover(Pop[0], Pop[2]);}

\hspace{0.5cm}\textsf{Pop[6] = Mutate(Pop[1]);}

\hspace{0.5cm}\textsf{Pop[2] = Select(Pop[4], Pop[3]);}

\hspace{0.5cm}\textsf{Pop[1] = Mutate(Pop[6]);}

\hspace{0.5cm}\textsf{Pop[3] = Crossover(Pop[5], Pop[1]);}

\textsf{... }

\textsf{{\}}}\\

These statements will be considered to be genetic operations executed 
during an EA generation. Since our purpose is to evolve a generational EA we 
have to add a wrapper loop around the genetic operations that are executed 
during an EA generation. More than that, each EA starts with a random 
population of individuals. Thus, the LGP program must contain some 
instructions that initialize the initial population. 

The obtained LGP chromosome is given below:\\

\textsf{\textbf{void}}\textsf{ 
LGP{\_}Program(}\textsf{\textbf{Chromosome}}
\textsf{ Pop[8]) // a population with of 8 individuals}

\textsf{{\{}}

\hspace{0.5cm}\textsf{Randomly{\_}initialize{\_}the{\_}population();}

\hspace{0.5cm}\textsf{\textbf{for}}\textsf{ (}\textsf{\textbf{int}}\textsf{ k = 0; k $<$ 
MaxGenerations; k++){\{} // repeat for a number of generations}

\hspace{1cm}\textsf{Pop[0] = Mutate(Pop[5]);}

\hspace{1cm}\textsf{Pop[7] = Select(Pop[3], Pop[6]);}

\hspace{1cm}\textsf{Pop[4] = Mutate(Pop[2]);}

\hspace{1cm}\textsf{Pop[2] = Crossover(Pop[0], Pop[2]);}

\hspace{1cm}\textsf{Pop[6] = Mutate(Pop[1]);}

\hspace{1cm}\textsf{Pop[2] = Select(Pop[4], Pop[3]);}

\hspace{1cm}\textsf{Pop[1] = Mutate(Pop[6]);}

\hspace{1cm}\textsf{Pop[3] = Crossover(Pop[5], Pop[1]);}

\hspace{1cm}\textsf{{\}}}

\textsf{{\}}}\\

\textbf{\textit{Remark}}: The initialization function and the \textbf{for} 
cycle will not be affected by the genetic operators. These parts are kept 
unchanged during the search process.

\subsection{Fitness Assignment}

We deal with EAs at two different levels: a micro level representing the 
evolutionary algorithm encoded into an LGP chromosome and a macro level GA, 
which evolves LGP individuals. Macro level GA execution is bounded by known 
rules for GAs (see (Goldberg, 1989)). 

In order to compute the fitness of a LGP individual we have to compute the quality 
of the EA encoded in that chromosome. For this purpose the EA encoded into a 
LGP chromosome is run on the particular problem being solved.

Roughly speaking the fitness of an LGP individual equals the fitness of 
the best solution generated by the evolutionary algorithm encoded into that 
LGP chromosome. But since the EA encoded into a LGP chromosome uses 
pseudo-random numbers it is very likely that successive runs of the same EA 
will generate completely different solutions. This stability problem is 
handled in a standard manner: the EA encoded into an LGP chromosome is 
executed (run) more times (500 runs are in fact executed in all the 
experiments performed for evolving EAs for function optimization and 25 runs 
for evolving EAs for TSP and QAP) and the fitness of a LGP chromosome is the 
average of the fitness of the EA encoded in that chromosome over all the runs.

The optimization type (minimization/maximization) of the macro level EA is the same as the optimization type of the micro level EA. In our experiments we have employed a minimization relation (finding the minimum of a function and finding the shortest TSP path and finding the minimal quadratic assignment).

\textit{Remark}. In standard LGP one of the registers is chosen 
as the program output. This register is not changed during the search 
process. In our approach the register storing the best value (best fitness) in all the generations
is chosen to represent the chromosome. Thus, every LGP chromosome stores 
multiple solutions of a problem in the same manner as Multi Expression 
Programming does (Oltean et al., 2003; Oltean, 2003; Oltean et al., 2004b).

\subsection{The Model used for Evolving EAs}

For evolving EAs we use the steady state algorithm described in section \ref{ss}. The problem set 
is divided into two sets, suggestively called training set, 
and test set. In our experiments the training set 
consists in a difficult test problem. The test set consists in some other well-known 
benchmarking problems (Burkard et al., 1991; Reinelt, 1991; Yao et al., 1999).

\section{Numerical Experiments}\label{sec4}

In this section several numerical experiments for evolving EAs are 
performed. Two evolutionary algorithms for function optimization, the TSP and the QAP 
problems are evolved. For assessing the performance of the evolved EAs, 
several numerical experiments with a standard Genetic Algorithm for function 
optimization, for TSP and for QAP are also performed and the results are compared.

\subsection{Evolving EAs for Function Optimization}\label{funcopt}

In this section an Evolutionary Algorithm for function optimization is evolved.

\subsubsection{Test Functions}

Ten test problems $f_{1}-f_{10}$ (given in Table \ref{tab1}) are used in order to asses the 
performance of the evolved EA. Functions $f_{1}-f_{6}$ are unimodal test 
function. Functions $f_{7}-f_{10}$ are highly multimodal (the number of the
local minima increases exponentially with the problem dimension (Yao et al., 1999)).

\begin{table}[htbp]
\caption{Test functions used in our experimental study. The parameter $n$ is the space dimension ($n$ = 5 in our numerical experiments) and $f_{min}$ is the minimum value of the function.}
\begin{center}
\begin{tabular}
{|p{200pt}|p{50pt}|p{50pt}|}
\hline
Test function& 
Domain& 
$f_{min}$ \\
\hline
$f_1 (x) = \sum\limits_{i = 1}^n {(i \cdot x_i^2 )} .$& 
[-10, 10]$^{ n}$& 
0 \\
\hline
$f_2 (x) = \sum\limits_{i = 1}^n {x_i^2 } .$& 
[-100, 100]$^{ n}$& 
0 \\
\hline
$f_3 (x) = \sum\limits_{i = 1}^n {\vert x_i \vert + \prod\limits_{i = 1}^n {\vert x_i \vert } } .$& 
[-10, 10]$^{ n}$& 
0 \\
\hline
$f_4 (x) = \sum\limits_{i = 1}^n {\left( {\sum\limits_{j = 1}^i {x_j^2 } } \right)} .$& 
[-100, 100$^{ }$]$^{ n}$& 
0 \\
\hline
$f_5 (x) = \max \{x_i ,1 \le i \le n\}.$& 
[-100, 100]$^{ n}$& 
0 \\
\hline
$f_{6} (x) = \sum\limits_{i = 1}^{n - 1} {100 \cdot (x_{i + 1} - x_i^2 )^2 + (1 - x_i )^2} .$& 
[-30, 30]$^{ n}$& 
0 \\
\hline
$f_7 (x) = 10 \cdot n + \sum\limits_{i = 1}^n {(x_i^2 - 10 \cdot \cos (2 \cdot \pi \cdot x_i ))} $& 
[-5, 5]$^{ n}$& 
0 \\
\hline
$f_8 (x) = - a \cdot e^{ - b\sqrt {\frac{\sum\limits_{i = 1}^n {x_i^2 } }{n}} } - e^{\frac{\sum {\cos (c \cdot x_i )} }{n}} + a + e.$& 
[-32, 32]$^{ n}$ \par $a$ = 20, $b$ = 0.2, $c$ = 2\textit{$\pi $}.& 
0 \\
\hline
$f_9 (x) = \frac{1}{4000} \cdot \sum\limits_{i = 1}^n {x_i^2 - \prod\limits_{i = 1}^n {\cos (\frac{x_i }{\sqrt i }) + 1} } .$& 
[-500, 500]$^{ n}$& 
0 \\
\hline
$f_{10} (x) = \sum\limits_{i = 1}^n {( - x_i \cdot \sin (\sqrt {\left| {x_i } \right|} ))} $& 
[-500, 500]$^{ n}$& 
-$n * $ 418.98 \\
\hline
\end{tabular}
\end{center}
\label{tab1}
\end{table}

\subsubsection{Experimental Results}\label{ga}

In this section we evolve an EA for function optimization and then we asses 
the performance of the evolved EA. A comparison with standard GA is 
performed farther in this section.

For evolving an EA we use $f_{1}$ as the training problem.

An important issue concerns the solutions evolved by the EAs encoded into an 
LGP chromosome and the specific genetic operators used for this purpose. 
The solutions evolved by the EA encoded into LGP chromosomes are represented 
using real values (Goldberg, 1989). Thus, each chromosome of the evolved EA is a fixed-length array of real values. By initialization, a point within the definition 
domain is randomly generated. Convex crossover with $\alpha $ = 
$\raise.5ex\hbox{$\scriptstyle 1$}\kern-.1em/ 
\kern-.15em\lower.25ex\hbox{$\scriptstyle 2$} $ and Gaussian mutation with 
$\sigma $ = 0.5 are used (Goldberg, 1989).

A short description of real encoding and the corresponding genetic operators is given in Table \ref{real}.

\begin{table}[htbp]
\caption{A short description of real encoding.}
\begin{center}
\begin{tabular}
{|p{115pt}|p{200pt}|}
\hline
Function to be optimized& 
$f$:[\textit{MinX}, \textit{MaxX}]$^{n} \to \Re $ \\
\hline
Individual representation& 
$x$ = ($x_{1}$, $x_{2}$, \ldots , $x_{n})$. \\
\hline
Convex Recombination with $\alpha $ = 0.5& 
parent 1 -- $x$ = ($x_{1}$, $x_{2}$, \ldots , $x_{n})$. \par parent 2 -- $y$ = ($y_{1}$, $y_{2}$, \ldots , $y_{n})$. \par the offspring -- $o = (\frac{x_1 + y_1 }{2},\,\frac{x_2 + y_2 }{2},\,...\,,\,\frac{x_n + y_n }{2}).$ \\
\hline
Gaussian Mutation& 
the parent -- $x$ = ($x_{1}$, $x_{2}$, \ldots , $x_{n})$. \par the offspring -- $o$ = ($x_{1 }+G$(0,$\sigma )$, $x_{2 }+G$(0,$\sigma )$, \ldots , $x_{n }+G$(0,$\sigma ))$, \par where $G$ is a function that generates real values with Gaussian distribution. \\
\hline
\end{tabular}
\end{center}
\label{real}
\end{table}

\textbf{Experiment 1}\\

In this experiment an Evolutionary Algorithm for function optimization is evolved.

There is a wide range of Evolutionary Algorithms that can be evolved by using the technique 
described above. Since the evolved EA has to be compared with another 
algorithm (such as standard GA or ES), the parameters of the evolved EA 
should be similar to the parameters of the algorithm used for comparison.

For instance, standard GA uses a primary population of $N$ individuals and an 
additional population (the new population) that stores the offspring 
obtained by crossover and mutation. Thus, the memory requirement for a 
standard GA is 2 * $N$. In each generation there will be 2 * $N$ Selections, $N$ 
Crossovers and $N$ Mutations (we assume here that only one offspring is 
obtained by the crossover of two parents). Thus, the number of genetic operators 
(\textit{Crossovers}, \textit{Mutations} and \textit{Selections}) in a standard GA is 4 * $N$. We do not take into account the complexity of the genetic operators, since in most of the cases this 
complexity is different from operator to operator. The standard GA algorithm 
is given below:

\begin{center}
\textbf{Standard GA algorithm}
\end{center}

\textsf{S}$_{1}$\textsf{. Randomly create the initial population 
}\textsf{\textit{P}}\textsf{(0)}

\textsf{S}$_{2}$\textsf{. }\textsf{\textbf{for}}\textsf{ 
}\textsf{\textit{t}}\textsf{ = 1 }\textsf{\textbf{to}}\textsf{ 
}\textsf{\textit{Max}}\textsf{{\_}}\textsf{\textit{Generations}}\textsf{ 
}\textsf{\textbf{do}}

\textsf{S}$_{3}$\textsf{. 
}\hspace{0.5cm}\textsf{\textit{P'}}\textsf{(}\textsf{\textit{t}}\textsf{) = $\phi $;}

\textsf{S}$_{4}$\textsf{.}\hspace{0.5cm}\textsf{\textbf{for}}\textsf{ 
}\textsf{\textit{k}}\textsf{ = 1 }\textsf{\textbf{to}}\textsf{ $\vert 
$}\textsf{\textit{P}}\textsf{(}\textsf{\textit{t}}\textsf{)$\vert $ 
}\textsf{\textbf{do}}

\textsf{S}$_{5}$\textsf{.}\hspace{1cm}\textsf{\textit{p}}$_{1}$\textsf{ = 
}\textsf{\textit{Select}}\textsf{(}\textsf{\textit{P}}\textsf{(}\textsf{\textit{t}}\textsf{)); 
// select an individual from the population}

\textsf{S}$_{6}$\textsf{.}\hspace{1cm}\textsf{\textit{p}}$_{2}$\textsf{ = 
}\textsf{\textit{Select}}\textsf{(}\textsf{\textit{P}}\textsf{(}\textsf{\textit{t}}\textsf{)); 
// select the second individual }

\textsf{S}$_{7}$\textsf{.}\hspace{1cm}\textsf{\textit{Crossover}}\textsf{ 
(}\textsf{\textit{p}}$_{1}$\textsf{, }\textsf{\textit{p}}$_{2}$\textsf{, 
}\textsf{\textit{offsp}}\textsf{); // crossover the parents p}$_{1}$\textsf{ 
and p}$_{2}$

\hspace{1.5cm}\textsf{// an offspring }\textsf{\textit{offspr}}\textsf{ is obtained}

\textsf{S}$_{8}$\textsf{.}\hspace{1cm}\textsf{\textit{Mutation}}\textsf{ 
(}\textsf{\textit{offspr}}\textsf{); // mutate the offspring 
}\textsf{\textit{offspr}}

\textsf{S}$_{9}$\textsf{.}\hspace{1cm}\textsf{ Add }\textsf{\textit{offspf}}\textsf{ to 
}\textsf{\textit{P'}}\textsf{(}\textsf{\textit{t}}\textsf{); //move 
}\textsf{\textit{offspr} in the new population}

\textsf{S}$_{10}$\textsf{.}\hspace{0.5cm}\textsf{\textbf{endfor}}

\textsf{S}$_{11}$\textsf{.}\hspace{0.5cm}\textsf{\textit{P}}\textsf{(}\textsf{\textit{t}}\textsf{+1) = 
}\textsf{\textit{P}}\textsf{'(}\textsf{\textit{t}}\textsf{);}

\textsf{S}$_{12}$\textsf{. }\textsf{\textbf{endfor}}\\

The best solution generated over all the generations is the output of the program.

Rewritten as an LGP program, the Standard GA is given below. The individuals 
of the standard (main) population are indexed from 0 to \textit{PopSize} - 1 and the 
individuals of the new population are indexed from \textit{PopSize} up to 2 * \textit{PopSize} - 1.\\

\textsf{\textbf{void}}\textsf{ 
}\textsf{\textit{LGP{\_}Program}}\textsf{(}\textsf{\textbf{Chromosome}}\textsf{ 
}\textsf{\textit{Pop}}\textsf{[2 * }\textsf{\textit{PopSize}}\textsf{]) }

\textsf{//an array containing of 2 * }\textsf{\textit{PopSize}}\textsf{ 
individuals}

\textsf{{\{}}

\textsf{Randomly{\_}initialize{\_}the{\_}population();}

\textsf{\textbf{for}}\textsf{ (}\textsf{\textbf{int}}\textsf{ $k$ = 0; $k <$ 
}\textsf{\textit{MaxGenerations}}\textsf{; $k$++){\{} // repeat for a number of generations}

\hspace{0.5cm}\textsf{// create the new population}

\hspace{0.5cm}\textsf{p1 = 
}\textsf{\textit{Select}}\textsf{(}\textsf{\textit{Pop}}\textsf{[1], 
}\textsf{\textit{Pop}}\textsf{[6]);}

\hspace{0.5cm}\textsf{p2 = 
}\textsf{\textit{Select}}\textsf{(}\textsf{\textit{Pop}}\textsf{[3], 
}\textsf{\textit{Pop}}\textsf{[2]);}

\hspace{0.5cm}\textsf{o = }\textsf{\textit{Crossover}}\textsf{(p1, p2);}

\hspace{0.5cm}\textsf{\textit{Pop}}\textsf{[}\textsf{\textit{PopSize}}\textsf{] = 
}\textsf{\textit{Mutate}}\textsf{(o);}\\

\hspace{0.5cm}\textsf{p1 = 
}\textsf{\textit{Select}}\textsf{(}\textsf{\textit{Pop}}\textsf{[3], 
}\textsf{\textit{Pop}}\textsf{[6]);}

\hspace{0.5cm}\textsf{p2 = 
}\textsf{\textit{Select}}\textsf{(}\textsf{\textit{Pop}}\textsf{[7], 
}\textsf{\textit{Pop}}\textsf{[1]);}

\hspace{0.5cm}\textsf{o = }\textsf{\textit{Crossover}}\textsf{(p1, p2);}

\hspace{0.5cm}\textsf{\textit{Pop}}\textsf{[}\textsf{\textit{PopSize}}\textsf{ + 1] = 
}\textsf{\textit{Mutate}}\textsf{(o);}\\

\hspace{0.5cm}\textsf{p1 = 
}\textsf{\textit{Select}}\textsf{(}\textsf{\textit{Pop}}\textsf{[2], 
}\textsf{\textit{Pop}}\textsf{[1]);}

\hspace{0.5cm}\textsf{p2 = 
}\textsf{\textit{Select}}\textsf{(}\textsf{\textit{Pop}}\textsf{[4], 
}\textsf{\textit{Pop}}\textsf{[7]);}

\hspace{0.5cm}\textsf{o = }\textsf{\textit{Crossover}}\textsf{(p1, p2);}

\hspace{0.5cm}\textsf{\textit{Pop}}\textsf{[}\textsf{\textit{PopSize}}\textsf{ + 2] = 
}\textsf{\textit{Mutate}}\textsf{(o);}

\hspace{0.5cm}\textsf{...}

\hspace{0.5cm}\textsf{p1 = 
}\textsf{\textit{Select}}\textsf{(}\textsf{\textit{Pop}}\textsf{[1], 
}\textsf{\textit{Pop}}\textsf{[5]);}

\hspace{0.5cm}\textsf{p2 = 
}\textsf{\textit{Select}}\textsf{(}\textsf{\textit{Pop}}\textsf{[7], 
}\textsf{\textit{Pop}}\textsf{[3]);}

\hspace{0.5cm}\textsf{o = }\textsf{\textit{Crossover}}\textsf{(p1, p2);}

\hspace{0.5cm}\textsf{\textit{Pop}}\textsf{[2 * }\textsf{\textit{PopSize}}\textsf{ - 1] = 
}\textsf{\textit{Mutate}}\textsf{(o);}\\

\hspace{0.5cm}\textsf{// pop(}\textsf{\textit{t}}\textsf{ + 1) = new{\_}pop 
(}\textsf{\textit{t}}\textsf{)}

\hspace{0.5cm}\textsf{// copy the individuals from new{\_}pop to the next population}\\

\hspace{0.5cm}\textsf{\textit{Pop}}\textsf{[0] = 
}\textsf{\textit{Pop}}\textsf{[}\textsf{\textit{PopSize}}\textsf{];}

\hspace{0.5cm}\textsf{\textit{Pop}}\textsf{[1] = 
}\textsf{\textit{Pop}}\textsf{[}\textsf{\textit{PopSize}}\textsf{ + 1];}

\hspace{0.5cm}\textsf{\textit{Pop}}\textsf{[2] = 
}\textsf{\textit{Pop}}\textsf{[}\textsf{\textit{PopSize}}\textsf{ + 2];}

\hspace{0.5cm}\textsf{...}

\hspace{0.5cm}\textsf{\textit{Pop}}\textsf{[}\textsf{\textit{PopSize}}\textsf{ - 1] = 
Pop[2 * }\textsf{\textit{PopSize}}\textsf{ - 1];}

\textsf{{\}}}

\textsf{{\}}}\\

The parameters of the standard GA are given in Table \ref{tab2}.

\begin{table}[htbp]
\caption{The parameters of a standard GA for Experiment 1.}
\begin{center}
\begin{tabular}
{|p{140pt}|p{175pt}|}
\hline
\textbf{Parameter}& 
\textbf{Value} \\
\hline
Population size& 
20 (+ 20 individuals in the new pop) \\
\hline
Individual encoding& 
fixed-length array of real values\\
\hline
Number of generations& 
100 \\
\hline
Crossover probability& 
1 \\
\hline
Crossover type& 
Convex Crossover with $\alpha $ = 0.5 \\
\hline
Mutation & 
Gaussian mutation with $\sigma $ = 0.01  \\
\hline
Mutation probability&
1\\
\hline
Selection& 
Binary Tournament \\
\hline
\end{tabular}
\end{center}
\label{tab2}
\end{table}

We will evolve an EA that uses the same memory requirements and the same 
number of genetic operations as the standard GA described above. \\

\textit{Remark}. We have performed several comparisons between the evolved EA and the standard GA. These comparisons are mainly based on two facts:

\begin{itemize}

\item[{\it (i)}]{the memory requirements (i.e. the population size) and the number of genetic operators used during 
the search process.}

\item[{\it (ii)}] the number of function evaluations. This comparison cannot be easily performed in our model since we 
cannot control the number of function evaluations (this number is decided by evolution). The total number of genetic operators (crossovers + mutations + selections) is the only parameter that can be controlled in our 
model. However, in order to perform a comparison based on the number of function evaluations we will adopt the following strategy: we will count the number function evaluations/generation performed by our evolved EA and we will use for comparison purposes another standard evolutionary algorithm (like GA) that performs the same number of function evaluations/generation. For instance, if our evolved EA performs 53 function evaluations/generation we will use a population of 53 individuals for the standard GA (knowing that the GA described in section \ref{ga} creates in each generation a number of new individuals equal to the population size).

\end{itemize}

The parameters of the LGP algorithm are given in Table \ref{tab3}.

\begin{table}[htbp]
\caption{The parameters of the LGP algorithm used for Experiment 1.}
\begin{center}
\begin{tabular}
{|p{140pt}|p{171pt}|}
\hline
\textbf{Parameter}& 
\textbf{Value} \\
\hline
Population size& 
500 \\
\hline
Code Length& 
80 instructions \\
\hline
Number of generations& 
100 \\
\hline
Crossover probability& 
0.7 \\
\hline
Crossover type& 
Uniform Crossover \\
\hline
Mutation & 
5 mutations per chromosome \\
\hline
Function set& 
$F$ = {\{}\textit{Select}, \textit{Crossover}, \textit{Mutate}{\}} \\
\hline
\end{tabular}
\end{center}
\label{tab3}
\end{table}

The parameters of the evolved EA are given in Table \ref{tab4}.

\begin{table}[htbp]
\caption{The parameters of the evolved EA for function optimization.}
\begin{center}
\begin{tabular}
{|p{140pt}|p{170pt}|}
\hline
\textbf{Parameter}& 
\textbf{Value} \\
\hline
Individual representation&
fixed-length array of real values.\\
\hline
Population size& 
40 \\
\hline
Number of generations& 
100 \\
\hline
Crossover probability& 
1 \\
\hline
Crossover type& 
Convex Crossover with $\alpha $ = 0.5 \\
\hline
Mutation & 
Gaussian mutation with $\sigma $ = 0.01  \\
\hline
Mutation probability&
1\\
\hline
Selection& 
Binary Tournament \\
\hline
\end{tabular}
\end{center}
\label{tab4}
\end{table}

\begin{figure}[htbp]
\centerline{\includegraphics[width=3.52in,height=3.16in]{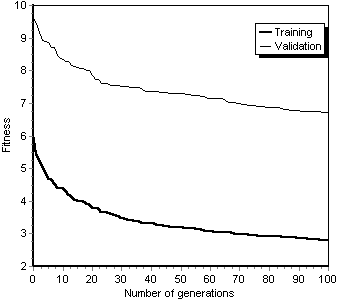}}
\caption{The relationship between the fitness of the best LGP individual in each generation and the number of generations. Results are averaged over 25 runs.}
\label{fig1}
\end{figure}

The results of this experiment are depicted in Figure \ref{fig1}.

The effectiveness of our approach can be seen in Figure \ref{fig1}. The LGP 
technique is able to evolve an EA for solving optimization problems. The 
quality of the evolved EA (LGP chromosome) improves as the search process advances.

\textbf{Experiment 2}\\

This experiment serves our purpose of comparing the evolved EA with the standard Genetic 
Algorithm described in Experiment 1. The parameters used by the evolved EA 
are given in Table \ref{tab4} and the parameters used by standard GA are given in 
Table \ref{tab2}. The results of the comparison are given in Table \ref{tab5}.

\begin{table}[htbp]
\caption{The results obtained by applying the Evolved EA and the Standard GA for the considered test functions. StdDev stands for the standard deviation. The results are averaged over 500 runs.}
\begin{center}
\begin{tabular}
{|p{55pt}|p{63pt}|p{63pt}|p{65pt}|p{65pt}|}
\hline
\raisebox{-1.50ex}[0cm][0cm]{\textbf{Test function }}& 
\multicolumn{2}{|p{127pt}|}{\textbf{Evolved EA} \par \textbf{40 individuals}} & 
\multicolumn{2}{|p{141pt}|}{\textbf{Standard GA} \par \textbf{20 individuals in the standard population + 20 individuals in the new population}}  \\
\cline{2-5} 
 & 
Mean& 
StdDev& 
Mean& 
StdDev \\
\hline
$f_{1}$& 
1.06& 
1.81& 
13.52& 
13.68 \\
\hline
$f_{2}$& 
104.00& 
115.00& 
733.40& 
645.80 \\
\hline
$f_{3}$& 
1.01& 
0.88& 
3.95& 
2.29 \\
\hline
$f_{4}$& 
149.02& 
163.37& 
756.61& 
701.52 \\
\hline
$f_{5}$& 
6.27& 
3.52& 
17.03& 
7.73 \\
\hline
$f_{6}$& 
2440.30& 
5112.72& 
113665.57& 
307109.69 \\
\hline
$f_{7}$& 
2.65& 
1.75& 
6.16& 
4.10 \\
\hline
$f_{8}$& 
5.08& 
2.34& 
10.39& 
2.90 \\
\hline
$f_{9}$& 
1.09& 
8.07& 
5.34& 
4.07 \\
\hline
$f_{10}$& 
-959.00& 
182.00& 
-860.39& 
202.19 \\
\hline
\end{tabular}
\end{center}
\label{tab5}
\end{table}

Table \ref{tab5} shows that the Evolved EA significantly outperforms 
the standard GA on all the considered test problems. 

The next experiment serves our purpose of comparing the Evolved EA with a Genetic Algorithm 
that performs the same number of function evaluations. Having this in view we count how many new individuals are created during a generation of the evolved EA. Thus, 
GA will use a main population of 56 individuals and a secondary 
population of 56 individuals. Note that this will provide significant advantage of the standard GA over the 
Evolved EA. However, we use this larger population because, in this case, 
the algorithms (the Standard GA and the Evolved EA) share an important 
parameter: they perform the same number of function evaluations. 
The results are presented in Table \ref{tab7}.

\begin{table}[htbp]
\caption{The results of applying the Evolved EA and the Standard GA for the considered test functions. StdDev stands for standard deviation. Results are averaged over 500 runs.}
\begin{center}
\begin{tabular}
{|p{60pt}|p{63pt}|p{63pt}|p{70pt}|p{70pt}|}
\hline
\raisebox{-1.50ex}[0cm][0cm]{\textbf{Test function }}& 
\multicolumn{2}{|p{127pt}|}{\textbf{Evolved EA} \par \textbf{40 individuals}} & 
\multicolumn{2}{|p{141pt}|}{\textbf{Standard GA} \par \textbf{56 individuals in the standard population + 56 individuals in the new population}}  \\
\cline{2-5} 
 & 
Mean& 
StdDev& 
Mean& 
StdDev \\
\hline
$f_{1}$& 
1.06& 
1.81& 
1.12& 
1.98 \\
\hline
$f_{2}$& 
104.00& 
115.00& 
90.10& 
108.02 \\
\hline
$f_{3}$& 
1.01& 
0.88& 
1.10& 
0.94 \\
\hline
$f_{4}$& 
149.02& 
163.37& 
111.09& 
128.01 \\
\hline
$f_{5}$& 
6.27& 
3.52& 
5.86& 
3.20 \\
\hline
$f_{6}$& 
2440.30& 
5112.72& 
2661.83& 
7592.10 \\
\hline
$f_{7}$& 
2.65& 
1.75& 
2.32& 
1.60 \\
\hline
$f_{8}$& 
5.08& 
2.34& 
5.08& 
2.24 \\
\hline
$f_{9}$& 
1.09& 
8.07& 
1.09& 
7.53 \\
\hline
$f_{10}$& 
-959.00& 
182.00& 
-1010.00& 
177.00 \\
\hline
\end{tabular}
\end{center}
\label{tab7}
\end{table}

The results in Table \ref{tab7} show that the Evolved EA is better than the standard 
GA in 3 cases (out of 10) and have the same average performance for 2 functions. However, in this case the standard GA has  
considerable advantage over the Evolved EA.\\

In order to determine whether the differences (given in Table \ref{tab5}) between the Evolved EA and the standard GA are statistically 
significant we use a $t$-test with 95{\%} confidence. Before applying the 
$t$-test, an $F$-test has been used for determining whether the compared data have the 
same variance. The $P$-values of a two-tailed $t$-test are given in Table \ref{tab_ttest}.

\begin{table}[htbp]
\caption{The results of the t-Test and F-Test.}
\begin{center}
\begin{tabular}
{|p{50pt}|p{54pt}|p{52pt}|}
\hline
Function& 
F-Test& 
t-Test \\
\hline
$f_{1}$& 
0.04& 
0.58 \\
\hline
$f_{2}$& 
0.17& 
0.05 \\
\hline
$f_{3}$& 
0.13& 
0.15 \\
\hline
$f_{4}$& 
7E-8& 
5E-5 \\
\hline
$f_{5}$& 
0.03& 
0.05 \\
\hline
$f_{6}$& 
1E-18& 
0.67 \\
\hline
$f_{7}$& 
0.04& 
1E-5 \\
\hline
$f_{8}$& 
0.32& 
0.99 \\
\hline
$f_{9}$& 
0.12& 
0.94 \\
\hline
$f_{10}$& 
0.47& 
2E-6 \\
\hline
\end{tabular}
\end{center}
\label{tab_ttest}
\end{table}

Table \ref{tab_ttest} shows that the difference between the Evolved EA and the 
standard GA is statistically significant ($P < 0.05$) for 3 test problems.\\

\textbf{Experiment 3}\\

We are also interested in analyzing the relationship between the number of 
generations of the evolved EA and the quality of the solutions obtained by 
applying the evolved EA for the considered test functions. The parameters of 
the Evolved EA (EEA) are given in Table \ref{tab4} and the parameters of the Standard 
GA (SGA) are given in Table \ref{tab2}. In order to provide a comparison based on the number of function evaluations we use a main population of 56 individuals for the Genetic Algorithm.

\begin{figure}[htbp]
\centerline{\includegraphics{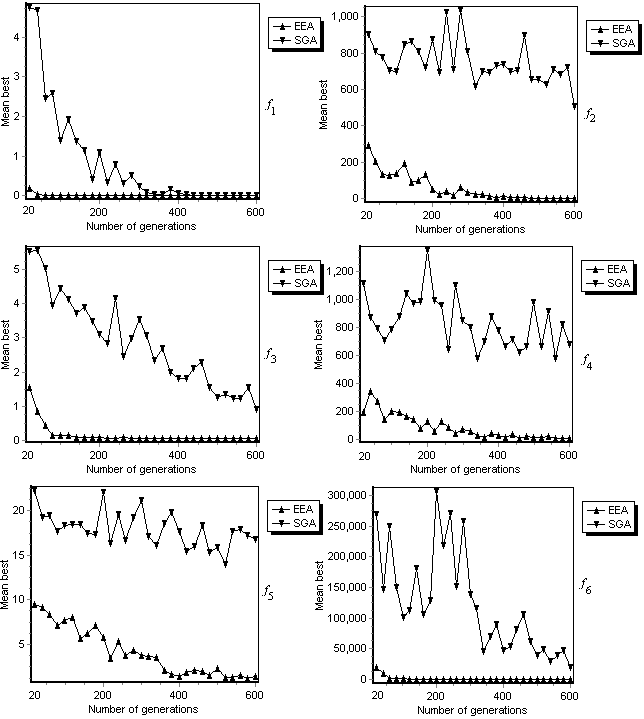}}
\caption{The relationship between the number of generations and the quality of the solutions obtained by the Evolved EA (EEA) and by the Standard GA (SGA) for the unimodal test functions $f_{1}-f_{6}$. The number of generations varies between 10 and 300. Results are averaged over 500 runs.}
\label{fig2}
\end{figure}

\begin{figure}[htbp]
\centerline{\includegraphics{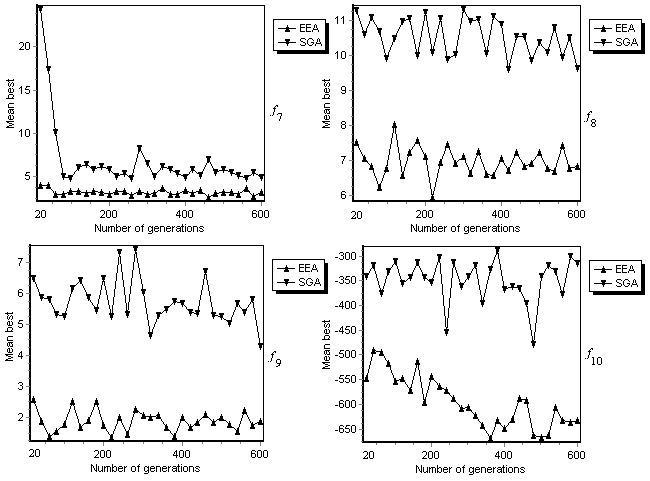}}
\caption{The relationship between the number of generations and the quality of the solutions obtained by the Evolved EA (EEA) and by the Standard GA (SGA) for the multimodal test functions $f_{7}-f_{10}$. The number of generations varies between 10 and 300. Results are averaged over 100 runs.}
\label{fig3}
\end{figure}

The results of this experiment are depicted in Figure \ref{fig2} (the unimodal 
test functions) and in Figure \ref{fig3} (the multimodal test functions).

Figures \ref{fig2} and \ref{fig3} show that the Evolved EA is scalable regarding the number of generations. For all test functions ($f_{1}-f_{10})$ we can see a continuous improvement tendency during the search process.

\subsection{Evolving EAs for TSP}\label{tsp}

In this section, an Evolutionary Algorithm for solving the Traveling 
Salesman Problem (Cormen et al., 1990; Garey et al., 1979) is evolved. First of all, the TSP problem is 
described and then an EA is evolved and its performance assessed by 
running it on several well-known instances in TSPLIB (Reinelt, 1991).\\

\subsubsection{The Traveling Salesman Problem}

The TSP may be stated as follows.

Consider a set $C$ = {\{}$c_{0}$, $c_{1}$,\ldots , $c_{N\mbox{--}1}${\}} of 
cities, and a distance $d(c_{i}$, $c_{j}) \quad  \in  \quad \Re ^{ + }$ for each pair 
$c_{i}$, $c_{j} \quad  \in  \quad C$. The tour $<$c$_{\pi (0)}$, c$_{\pi (1)}$, \ldots , 
c$_{\pi (N\mbox{-}1)}>$ of all cities in $C$ having minimum length is needed 
(Cormen et al., 1990; Garey et al., 1979).

The TSP is NP-complete (Garey et al., 1979). No polynomial time algorithm for 
solving this problem is known. Evolutionary Algorithms have been extensively 
used for solving this problem (Freisleben et al., 1996; Krasnogor, 2002; Merz et al., 1997).\\

\textbf{Experiment 5}\\

In this experiment, an EA for the TSP problem is evolved.

A TSP path will be represented as a permutation of cities (Freisleben et al., 1996; Merz et al., 1997) and it is 
initialized by using the Nearest Neighbor heuristic (Cormen et al., 1990; Garey et al., 1979). The genetic 
operators used by the Evolved EA are DPX as crossover and 2-Exchange (Krasnogor, 2002) as 
mutation. These operators are briefly described in what follows.

The DPX recombination operator copies into offspring all the common edges of the 
parents. Then it completes the offspring to achieve a valid tour with links 
that do not belong to the parents, in such a way that the distance between 
the parents in the newly created offspring is preserved. This completion may be 
done by using the nearest neighbor information (Freisleben et al., 1996; Merz et al., 1997).

Mutation is performed by applying 2-Exchange operator. The 2-\textit{Exchange} operator breaks the 
tour by 2 edges and then rebuilds the path by adding 2 new edges (see (Krasnogor, 2002)).

The parameters used by the LGP algorithm are given in Table \ref{tab8}.

\begin{table}[htbp]
\caption{The parameters of the LGP algorithm used for Experiment 5.}
\begin{center}
\begin{tabular}
{|p{140pt}|p{171pt}|}
\hline
\textbf{Parameter}& 
\textbf{Value} \\
\hline
Population size& 
500 \\
\hline
Code Length& 
80 instructions \\
\hline
Number of generations& 
50 \\
\hline
Crossover probability& 
0.7 \\
\hline
Crossover type& 
Uniform Crossover \\
\hline
Mutation & 
5 mutations per chromosome \\
\hline
Function set& 
$F$ = {\{}\textit{Select}, \textit{Crossover}, \textit{Mutate}{\}} \\
\hline
\end{tabular}
\end{center}
\label{tab8}
\end{table}

The parameters of the Evolved EA are given in Table \ref{tab9}.

\begin{table}[htbp]
\caption{The parameters of the evolved EA for TSP.}
\begin{center}
\begin{tabular}
{|p{140pt}|p{170pt}|}
\hline
\textbf{Parameter}& 
\textbf{Value} \\
\hline
Population size& 
40 \\
\hline
Number of generations& 
100 \\
\hline
Crossover probability& 
1 \\
\hline
Crossover type& 
DPX \\
\hline
Mutation & 
2-Exchange \\
\hline
Selection& 
Binary Tournament \\
\hline
\end{tabular}
\end{center}
\label{tab9}
\end{table}

For the training and testing stages of our algorithm we use several problems 
from the TSPLIB (Reinelt, 1991). The \textit{att48} problem (containing 48 nodes) is used for 
training purposes. Some other 25 well-known TSP instances are used as the test 
set.

25 runs for evolving EAs were performed. The time needed for a run was about a day on a 
PIII -600 MHz computer. An EA yielding a very good performance was 
evolved in each run. One of these EAs was tested against other 26 difficult 
instances from TSPLIB. 

The results of the Evolved EA along with the results 
obtained by using the GA described in section \ref{ga} are given in Table 
\ref{tab10}. Again we count the number of the newly created individuals in a generation of the evolved EA. Thus the standard GA will use a main population of 55 individuals and a secondary population of 55 individuals. In this way both algorithms will perform the same number of function evaluations.

\begin{table}[htbp]
\caption{The results of the standard GA and Evolved EA for 27 instances from TSPLIB. \textit{Mean} stands for the mean over all runs and \textit{StdDev} stands for the standard deviation. The difference $\Delta $ is in percent and it is computed considering the values of the Evolved EA as a baseline. Results are averaged over 100 runs.}
\begin{center}
\begin{tabular}
{|p{54pt}|p{58pt}|p{58pt}|p{63pt}|p{54pt}|p{36pt}|}
\hline
\raisebox{-1.50ex}[0cm][0cm]{\textbf{Problem}}& 
\multicolumn{2}{|p{117pt}|}{\textbf{Standard GA}} & 
\multicolumn{2}{|p{117pt}|}{\textbf{Evolved EA}} & 
\raisebox{-1.50ex}[0cm][0cm]{$\Delta $} \\
\cline{2-5} 
 & 
\textit{Mean}& 
\textit{StdDev}& 
\textit{Mean}& 
\textit{StdDev}& 
  \\
\hline
a280& 
3143.64& 
20.91& 
3051.35& 
39.34& 
3.02 \\
\hline
att48& 
37173.41& 
656.13& 
36011.50& 
650.19& 
3.22 \\
\hline
berlin52& 
8202.10& 
83.5758& 
7989.63& 
114.98& 
2.65 \\
\hline
bier127& 
127401.70& 
1119.56& 
126914.50& 
1295.47& 
0.38 \\
\hline
ch130& 
7124.14& 
86.98& 
6734.12& 
114.05& 
5.79 \\
\hline
ch150& 
7089.56& 
17.68& 
6950.81& 
97.36& 
1.99 \\
\hline
d198& 
17578.45& 
200.50& 
17127.13& 
220.12& 
2.63 \\
\hline
d493& 
40435.86& 
408.1137& 
39631.29& 
407.5115& 
2.03 \\
\hline
d657& 
59638.29& 
503.0018& 
58026.19& 
591.9782& 
2.77 \\
\hline
eil101& 
741.91& 
5.12& 
728.58& 
7.92& 
1.82 \\
\hline
eil51& 
468.91& 
5.06& 
461.25& 
4.22& 
1.66 \\
\hline
eil76& 
604.31& 
8.08& 
587.57& 
6.82& 
2.84 \\
\hline
fl417& 
14535.32& 
223.36& 
14288.14& 
198.99& 
1.72 \\
\hline
gil262& 
2799.96& 
26.56& 
2721.58& 
37.55& 
2.87 \\
\hline
kroA100& 
24496.34& 
235.40& 
23780.42& 
435.99& 
3.01 \\
\hline
kroA150& 
31690.82& 
374.15& 
30247.58& 
461.91& 
4.77 \\
\hline
kroA200& 
34647.90& 
278.16& 
33613.78& 
664.49& 
3.07 \\
\hline
kroB100& 
24805.07& 
281.13& 
23623.80& 
320.46& 
5.00 \\
\hline
kroB150& 
30714.54& 
425.03& 
29628.97& 
465.14& 
3.66 \\
\hline
kroC100& 
23328.12& 
336.89& 
22185.22& 
402.76& 
5.15 \\
\hline
kroD100& 
24716.68& 
195.48& 
24192.46& 
282.24& 
2.16 \\
\hline
kroE100& 
24930.71& 
202.08& 
24184.65& 
470.68& 
3.08 \\
\hline
lin105& 
16937.03& 
104.73& 
16324.81& 
432.41& 
3.75 \\
\hline
lin318& 
49813.96& 
454.12& 
49496.42& 
590.41& 
0.64 \\
\hline
p654& 
42827.31& 
522.99& 
40853.26& 
1004.75& 
4.83 \\
\hline
pcb442& 
59509.64& 
251.36& 
58638.23& 
485.51& 
1.48 \\
\hline
pr107& 
46996.24& 
362.04& 
46175.80& 
240.38& 
1.77 \\
\hline
\end{tabular}
\end{center}
\label{tab10}
\end{table}

From Table \ref{tab10} it can be seen that the Evolved EA performs better than the 
standard GA for all the considered test problems. The difference $\Delta $ 
ranges from 0.38 {\%} (for the problem \textit{bier127}) up to 5.79 {\%} (for the problem 
\textit{ch130}).

One can see that the standard GA performs very poorly compared with other 
implementations found in literature (Krasnogor, 2002; Merz et al., 1997). This is due to the weak 
(non-elitist) evolutionary scheme employed in this experiment. The 
performance of the GA can be improved by preserving the best individual 
found so far. We also could use the Lin-Kernighan heuristic (Freisleben et al., 1996; Merz et al., 1997) for genereting very initial solutions and thus improving the search. However, this is beyond the purpose of this research. Our main aim was to evolve an Evolutionary Algorithm and then to compare it with some 
similar (in terms of the number of genetic operations performed, the memory requirements and the number of function evaluations) EA structures.\\

\subsection{Evolving EAs for the Quadratic Assignment Problem}\label{qap}

In this section an evolutionary algorithm for the Quadratic Assignment Problem is evolved.

\subsubsection{The Quadratic Assignment Problem}

In the quadratic assignment problem (QAP), $n$ facilities have to be assigned 
to $n$ locations at the minimum cost. Given the set $\Pi $($n$) of all the permutations 
of {\{}1, 2, 3,\ldots $n${\}} and two $n$x$n$ matrices $A$=($a_{ij})$ and 
$B$=($b_{ij})$ the task is to minimize the quantity

\[
C(\pi ) = \sum\limits_{i = 1}^n {\sum\limits_{j = 1}^n {a_{ij} \cdot b_{\pi 
(i)\pi (j).\,\,\,\,\,\,\,\,\,\,\,\pi \in \Pi (n).} } } 
\]

Matrix $A$ can be interpreted as a distance matrix, i.e. $a_{ij}$ denotes the 
distance between location $i$ and location $j$, and $B$ is referred to as the flow 
matrix, i.e. $b_{kl}$ represents the flow of materials from facility $k$ to 
facility $l$. The QAP belongs to the class of NP-hard problems (Garey et al., 1979).\\

\textbf{Experiment 12}\\

In this experiment, an Evolutionary Algorithm for the QAP problem is 
evolved.

Every QAP solution is a permutation $\pi $ encoded as a vector of 
facilities, so that the value $j$ of the $i^{th}$ component in the vector 
indicates that the facility $j$ is assigned to location $i$ ($\pi (i)=j)$.

The initial population contains randomly generated individuals. The crossover 
operator is DPX (Merz et al., 2000). Mutation is performed by swapping two randomly chosen facilities (Merz et al., 2000).

The parameters used by the LGP algorithm are given in Table \ref{tab12}.

\begin{table}[htbp]
\caption{The parameters of the LGP algorithm used for evolving an Evolutionary Algorithm for the Quadratic Assignment Problem.}
\begin{center}
\begin{tabular}
{|p{140pt}|p{171pt}|}
\hline
\textbf{Parameter}& 
\textbf{Value} \\
\hline
Population size& 
500 \\
\hline
Code Length& 
80 instructions \\
\hline
Number of generations& 
50 \\
\hline
Crossover probability& 
0.7 \\
\hline
Crossover type& 
Uniform Crossover \\
\hline
Mutation & 
5 mutations per chromosome \\
\hline
Function set& 
$F$ = {\{}\textit{Select}, \textit{Crossover}, \textit{Mutate}{\}} \\
\hline
\end{tabular}
\end{center}
\label{tab12}
\end{table}

The parameters of the Evolved EA for QAP are given in Table \ref{tab13}.

\begin{table}[htbp]
\caption{The parameters of the evolved EA for the QAP.}
\begin{center}
\begin{tabular}
{|p{140pt}|p{189pt}|}
\hline
\textbf{Parameter}& 
\textbf{Value} \\
\hline
Population size& 
40 \\
\hline
Number of generations& 
100 \\
\hline
Crossover probability& 
1 \\
\hline
Crossover type& 
DPX \\
\hline
Mutation & 
2-Exchange \\
\hline
Selection& 
Binary Tournament \\
\hline
\end{tabular}
\end{center}
\label{tab13}
\end{table}

For the training and testing stages of our algorithm we use several problems 
from the QAPLIB (Burkard et al., 1991). The \textit{tai10a} problem (containing 10 facilities) is used for 
training purposes. Some other 26 well-known QAP instances are used as the test 
set.

25 runs for evolving EAs were performed. In each run an EA yielding a very 
good performance has been evolved. One of the evolved EAs was tested against other 26 difficult instances from QAPLIB. The 
results of the Evolved EA along with the results obtained with the GA 
described in section \ref{ga} are given in Table \ref{tab14}. Since the evolved EA 
creates 42 individuals at each generation we will use for the standard GA a 
main population of 42 individuals and an additional population (the new population) 
with 42 individuals.

\begin{table}[htbp]
\caption{The results of the standard GA and of the Evolved EA for 27 
instances from QAPLIB. \textit{Mean} stands for the mean over all runs and \textit{StdDev} stands for 
the standard deviation. The difference $\Delta $ is shown as a percentage and it is 
computed considering the values of the Evolved EA as a baseline. Results are 
averaged over 100 runs.}
\begin{center}
\begin{tabular}
{|p{60pt}|p{70pt}|p{50pt}|p{60pt}|p{50pt}|p{30pt}|}
\hline
\raisebox{-1.50ex}[0cm][0cm]{Problem}& 
\multicolumn{2}{|p{130pt}|}{\textbf{Standard GA}} & 
\multicolumn{2}{|p{130pt}|}{\textbf{Evolved EA}} & 
\raisebox{-1.50ex}[0cm][0cm]{$\Delta $} \\
\cline{2-5} 
 & 
\textit{Mean}& 
\textit{StdDev}& 
\textit{Mean}& 
\textit{StdDev}& 
  \\
\hline
bur26a& 
5496026.50& 
12150.28& 
5470251.89& 
14547.86& 
0.47 \\
\hline
chr12a& 
13841.42& 
1291.66& 
12288.16& 
1545.37& 
12.64 \\
\hline
chr15a& 
18781.02& 
1820.54& 
15280.78& 
1775.68& 
22.90 \\
\hline
chr25a& 
9224.26& 
600.94& 
7514.98& 
731.41& 
22.74 \\
\hline
esc16a& 
71.66& 
2.44& 
68.56& 
1.10& 
4.52 \\
\hline
had12& 
1682.10& 
9.40& 
1666.42& 
9.42& 
0.94 \\
\hline
had20& 
7111.54& 
46.19& 
7044.46& 
56.35& 
0.95 \\
\hline
kra30a& 
107165.20& 
1730.81& 
103566.40& 
2083.24& 
3.47 \\
\hline
kra32& 
21384.06& 
368.69& 
20727.38& 
374.24& 
3.16 \\
\hline
lipa50a& 
63348& 
48.02& 
63306.37& 
42.38& 
0.06 \\
\hline
nug30& 
6902.26& 
87.82& 
6774.00& 
85.40& 
1.89 \\
\hline
rou20& 
798047.00& 
7923.00& 
774706.80& 
8947.67& 
3.01 \\
\hline
scr20& 
141237.72& 
4226.88& 
128670.50& 
6012.24& 
9.76 \\
\hline
sko42& 
17684.18& 
159.31& 
17569.20& 
171.43& 
0.65 \\
\hline
sko49& 
25968.34& 
195.09& 
25895.76& 
186.31& 
0.28 \\
\hline
ste36a& 
13562.82& 
377.62& 
13022.90& 
457.48& 
4.14 \\
\hline
ste36b& 
31875.52& 
2095.00& 
29276.02& 
2254.04& 
8.87 \\
\hline
ste36c& 
11282157.00& 
320870.20& 
10899290.06& 
432078.30& 
3.51 \\
\hline
tai20a& 
785232.10& 
6882.52& 
759370.20& 
7808.40& 
3.40 \\
\hline
tai25a& 
1282398.50& 
7938.85& 
1256943.80& 
9985.62& 
2.02 \\
\hline
tai30a& 
2010495.90& 
14351.86& 
1978437.90& 
14664.52& 
1.62 \\
\hline
tai35a& 
2688498.90& 
17643.60& 
2649634.68& 
19598.61& 
1.46 \\
\hline
tai50a& 
5485928.90& 
29697.00& 
5461181.02& 
28383.97& 
0.45 \\
\hline
tai60a& 
7977368.30& 
35081.48& 
7960123.48& 
38001.33& 
0.21 \\
\hline
tho30& 
172923.82& 
2326.60& 
168152.84& 
2722.36& 
2.83 \\
\hline
tho40& 
281015.00& 
3890.10& 
277275.46& 
3555.32& 
1.34 \\
\hline
wil50& 
51751.84& 
250.69& 
51740.46& 
269.39& 
0.02 \\
\hline
\end{tabular}
\end{center}
\label{tab14}
\end{table}

Table \ref{tab14} shows that the evolved EA performs better than the standard GA for all the considered QAP instances. The difference $\Delta$ ranges from 0.02 (for the $wil50$ problem) up to 22.90 (for the $chr15a$ problem).

We could use some local search (Merz et al., 2000) techniques in order to improve the quality of the solutions, but this is again beyond the purpose of our research.

\section{Further work}\label{future}

Some other questions should be answered about the Evolved 
Evolutionary Algorithms. Some of them are:

\begin{itemize}
\item Are there patterns in the source code of the Evolved EAs? i.e. should we 
expect that the best algorithm for a given problem contain a patterned 
sequence of instructions? When a standard GA is concerned the sequence is the following: selection, recombination and mutations. Such a sequence has 
been given in section \ref{ga}. But, how do we know what the optimal sequence 
of instructions for a given problem is?

\item Are all the instructions effective? It is possible that a genetic operation 
be useless (i.e. two consecutive crossovers operating on the same two 
parents). Brameier and Banzhaf (Brameier et al., 2001a) used an algorithm that removes the 
introns from the LGP chromosomes. Unfortunately, this choice proved to be not 
very efficient in practice since some useless genetic material should be kept 
in order to provide a minimum of genetic diversity.

\item Are all the genetic operators suitable for the particular problem being 
solved? A careful analysis regarding the genetic operators used should be 
performed in order to obtain the best results. The usefulness / useless of the 
genetic operators employed by the GP has already been subject to long 
debates. Due to the NFL theorems (Wolpert et al., 1997) we know that we cannot have 
"the best" genetic operator that performs the best for all the problems. 
However, this is not our case, since our purpose is to find Evolutionary 
Algorithms for particular classes of problems.

\item What is the optimal number of genetic instructions performed during a 
generation of the Evolved EA? In the experiments performed in this paper we 
used fixed length LGP chromosomes. In this way we forced a certain number of 
genetic operations to be performed during a generation of the Evolved EA. 
Further numerical experiments will be performed by using variable length LGP 
chromosomes, hoping that this representation will find the optimal number 
of genetic instructions that have to be performed during a generation.

\end{itemize}

Another approach to the problem of evolving EAs could be based on Automatically Defined Functions (Koza, 1994). Instead of evolving an entire EA we will try to evolve a small pattern (sequence of instructions) that will be repeatedly used to generate new individuals. Most of the known evolutionary schemes use this form of evolution. For instance the pattern employed by a Genetic Algorithm is:\\

$p_1$ = Select ($Pop$[3], $Pop$[7]); // two individuals randomly chosen \\

$p_2$ = Select ($Pop$[5], $Pop$[1]); // another two individuals randomly chosen\\

$c$ = Crossover($p_1$, $p_2$);\\

$c$ = Mutate($c$);\\

An advantage of this approach is its reduced complexity: the size of the pattern is considerably smaller than the size of the entire EA.

In order to evolve high high quality EAs and assess their performance an 
extended set of training problems should be used. This set should include 
problems from different fields such as: function optimization, symbolic 
regression, TSP, classification etc. Further efforts will be dedicated to 
the training of such algorithm which should to have an increased generalization 
ability.

For obtaining more powerful Evolutionary Algorithms an extended set of operators will be used. This set will include operators that compute the fitness of the best/worst individual in the population. In this case the evolved EA will have the "elitism" feature which will allow us to compare it with more complex evolutionary schemes like steady-state (Syswerda, 1989).

In our experiments only populations of fixed size have been used. Another 
extension of the proposed approach will take into account the scalability of 
the population size.

Further numerical experiments will analyze the relationship between the LGP 
parameters (such as \textit{Population Size}, \textit{Chromosome Length}, \textit{Mutation Probability} etc.) and the ability of the evolved EA to find optimal solutions.

\section{Conclusions}

In this paper, Linear Genetic Programming has been used for evolving 
Evolutionary Algorithms. A detailed description of the proposed approach has 
been given allowing researchers to apply the method for evolving 
Evolutionary Algorithms that could be used for solving problems in their 
fields of interest. 

The proposed model has been used for evolving Evolutionary Algorithms for function optimization, the Traveling Salesman Problem and the Quadratic Assignment Problem. Numerical experiments emphasize the robustness and the 
efficacy of this approach. The evolved Evolutionary Algorithms perform similar and sometimes even better than some standard approaches in the literature.

\section*{Acknowledgments} 

The author likes to thanks anonymous reviewers for their usefull sugestions.

The source code for evolving Evolutionary Algorithms and all the evolved EAs described in this paper are available at www.eea.cs.ubbcluj.ro.


\begin{thebibliography}{99} 

\bibitem[]{angeline1}
Angeline, P.~J., (1995), Adaptive and Self-Adaptive Evolutionary Computations,  {\em Computational Intelligence: A Dynamic Systems Perspective}, pages 152-163,  IEEE Press: New York, USA.

\bibitem[]{angeline2}
Angeline, P.~J., (1996), Two Self-Adaptive Crossover Operators for Genetic Programming. In Angeline, P. and Kinnear, K.~E., editors, {\em Advances in Genetic Programming II}, pages 89-110, MIT Press, Cambridge, MA, USA.

\bibitem[]{back1}
Back, T. (1992), Self-Adaptation in Genetic Algorithms, In {\em Toward a Practice of Autonomous Systems: Proceedings of the first European Conference on Artificial Life}, pages 263-271, MIT Press, Cambridge, MA, USA.

\bibitem[]{banzhaf1}
Banzhaf, W., Nordin, P., Keller, R.~E. and Francone, F.~D. (1998). {\em Genetic Programming - An Introduction On the automatic evolution of computer programs and its applications}, dpunkt/Morgan Kaufmann, Heidelberg/San Francisco.

\bibitem[]{brameier1}
Brameier, M. and Banzhaf, W. (2001a). A Comparison of Linear Genetic Programming and 
Neural Networks in Medical Data Mining, {\em IEEE Transactions on Evolutionary 
Computation}, 5:17-26, IEEE Press, NY, USA.

\bibitem[]{brameier3}
Brameier, M. and Banzhaf, W. (2001b). Evolving Teams of Predictors with Linear Genetic 
Programming, {\em Genetic Programming and Evolvable Machines}, 2:381-407, Kluwer.

\bibitem[]{brameier2}
Brameier, M. and Banzhaf, W. (2002). Explicit Control of Diversity and Effective 
Variation Distance in Linear Genetic Programming, In E. Lutton, J. Foster, 
J. Miller, C. Ryan and A. Tettamanzi Editors, {\em European Conference on Genetic Programming IV}, Springer Verlag, Berlin, pages 38-50, 2002.

\bibitem[]{burkard1}
Burkard, R.~E. and Rendl, F. (1991), QAPLIB-A Quadratic Assignment Problem Libray, {\em European Journal of Operational Research}, 115-119.

\bibitem[]{cormen1}
Cormen, T.~H., Leiserson, C.~E. and Rivest, R.~R. (1990). {\em Introduction to Algorithms}, 
MIT Press.

\bibitem[]{edmonds1}
Edmonds, B. (2001). Meta-Genetic Programming: Co-evolving the Operators of 
Variation. {\em Electrik on AI}, 9:13-29.

\bibitem[]{freisleben1}
Freisleben, B. and Merz, P. (1996). A Genetic Local Search Algorithm for Solving 
Symmetric and Asymmetric Traveling Salesman Problems, In {\em IEEE International Conference on Evolutionary Computation 1996}, pages 616-621, IEEE Press.

\bibitem[]{garey1}
Garey, M.~R., and Johnson, D.~S. (1979). {\em Computers and Intractability: A Guide to 
NP-Completeness}, Freeman {\&} Co, San Francisco, CA.

\bibitem[]{goldberg1}
Goldberg, D.~E. (1989). {\em Genetic Algorithms in Search, Optimization, and Machine 
Learning}, Addison-Wesley, Reading, MA.

\bibitem[]{holland1}
Holland, J.~H. (1975). {\em Adaptation in Natural and Artificial Systems}, University of 
Michigan Press, Ann Arbor.

\bibitem[]{koza1}
Koza, J.~R. (1992). {\em Genetic Programming, On the Programming of Computers by Means of 
Natural Selection}, MIT Press, Cambridge, MA.

\bibitem[]{koza2}
Koza, J.~R. (1994). {\em Genetic Programming II, Automatic Discovery of Reusable 
Subprograms}, MIT Press, Cambridge, MA

\bibitem[]{krasnogor1}
Krasnogor, N. (2002) Studies on the Theory and Design Space of Memetic Algorithms, 
PhD Thesis, University of the West of England, Bristol, 2002.

\bibitem[]{merz1}
Merz, P. and Freisleben B. (1997). Genetic Local Search for the TSP: New Results, In 
{\em IEEE International Conference on Evolutionary Computation}, pages 616-621.

\bibitem[]{merz2}
Merz, P. and Freisleben, B. (2000), Fitness Landscape Analysis and Memetic Algorithms for the Quadratic Assignment Problem, {\em IEEE Transaction On Evolutionary Computation}, 4:337-352, IEEE Press, NY, USA.

\bibitem[]{nordin1}
Nordin, P. (1994). A Compiling Genetic Programming System that Directly Manipulates the
Machine-Code. In Kinnear K.E. editors, {\em Advances in Genetic Programming I}, pages 311-331,
MIT Press, Cambridge, MA, USA.

\bibitem[]{oltean1}
Oltean, M. and Gro\c san, C. (2003). Evolving Evolutionary Algorithms using Multi 
Expression Programming. In Banzhaf, W. (et al.) editors, {\em European Conference on Artificial Life VII}, LNAI 2801, pages 651-658, Springer-Verlag, Berlin, Germany.

\bibitem[]{oltean2}
Oltean, M. (2003). Solving Even-parity problems with Multi Expression Programming. 
In Chen K. (et al.) editors, {\em The $5^{th}$ International Workshop on Frontiers in Evolutionary Algorithm}, 
pages 315-318. 

\bibitem[]{oltean4}
Oltean, M. and Dumitrescu, D. (2004a). Evolving TSP heuristics with Multi Expression Programming, In Encoding Multiple Solutions in a Linear GP Chromosome. In Bubak, M., van Albada, G. D., Sloot, P., and Dongarra, J., {\em International Conference on Computational Sciences}, Vol II, pages 670-673, Springer-Verlag, Berlin.

\bibitem[]{oltean5}
Oltean, M., Gro\c san, C. and Oltean, M., (2004b). Encoding Multiple Solutions in a Linear GP Chromosome. In Bubak, M., van Albada, G. D., Sloot, P., and Dongarra, J., {\em International Conference on Computational Sciences, E-HARD Workshop}, Vol III, pages 1281-1288, Springer-Verlag, Berlin.

\bibitem[]{reinelt1}
Reinelt, G. (1991). TSPLIB - A Traveling Salesman Problem Library, ORSA, {\em Journal of 
Computing}, 4:376-384.

\bibitem[]{ross1}
Ross, B. (2002), Searching for Search Algorithms: Experiments in Meta-search, Technical Report: CS-02-23, Brock University, Ontario, Canada.

\bibitem[]{spector1}
Spector, L. and Robinson, A. (2002). Genetic Programming and Autoconstructive 
Evolution with the Push Programming Language. Genetic Programming and 
Evolvable Machines, 1:7-40, Kluwer.

\bibitem[]{stephens1}
Stephens, C.~R., Olmedo, I.~G., Vargas, J.~M. and Waelbroeck, H. (1998), Self-Adaptation in Evolving Systems, {\em Artificial Life} 4:183-201.

\bibitem[]{syswerda1}
Syswerda, G. (1989). Uniform Crossover in Genetic Algorithms. In Schaffer, J.~D. editors, {\em The 
3$^{rd}$ International Conference on Genetic Algorithms}, pages 2-9, Morgan Kaufmann Publishers, San Mateo, CA.

\bibitem[]{tavares1}
Tavares, J., Machado, P., Cardoso A., Pereira F.~B., Costa E. (2004). On the Evolution of Evolutionary Algorithms. In Keijzer, M. (et al.) editors, European Conference on Genetic Programming, pages 389-398, Springer-Verlag, Berlin.

\bibitem[]{teller1}
Teller, A. (1996). Evolving Programmers: the Co-evolution of Intelligent 
Recombination Operators. In Angeline, P. and Kinnear, K.~E., editors, {\em Advances in Genetic Programming II}, pages 45-68, MIT Press, USA.

\bibitem[]{yao1}
Yao, X., Liu, Y. and Lin, G. (1999). Evolutionary Programming Made Faster. {\em IEEE 
Transaction on Evolutionary Computation}, 2:82-102, IEEE Press, NY, USA.

\bibitem[]{wolpert1}
Wolpert, D.~H. and McReady, W.~G. (1995). No Free Lunch Theorems for Search, Technical Report SFI-TR-05-010, Santa Fe Institute, USA.

\bibitem[]{wolpert2}
Wolpert, D.~H. and McReady, W.~G. (1997). No Free Lunch Theorems for Optimization. {\em IEEE 
Transaction on Evolutionary Computation}, 1:67-82, IEEE Press, NY, USA.

\end{thebibliography}
\end{document}